\documentclass[twocolumn]{article}
\usepackage{spconf,amsmath,amsfonts, graphicx}
\usepackage{booktabs}
\usepackage{lipsum}  
\usepackage{subcaption}
\usepackage{tikz}
\usepackage{pifont}
\usepackage{tcolorbox}
\usepackage{enumitem}
\setlist{nolistsep}

\newcommand{\figTeaser}{
\begin{figure}[t]
    \centering
     \begin{subfigure}[b]{0.2\textwidth}
         \centering
         \includegraphics[width=\textwidth]{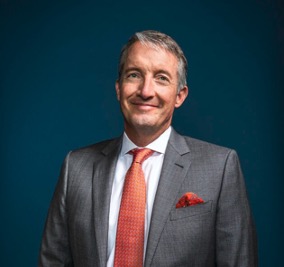}
     \end{subfigure}
     \begin{subfigure}[b]{0.2\textwidth}
         \centering
         \includegraphics[width=\textwidth]{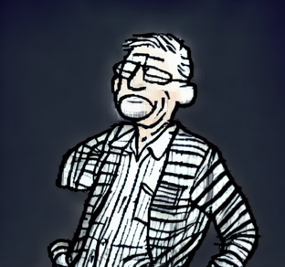}
     \end{subfigure}
     \hfill
     \begin{subfigure}[b]{0.2\textwidth}
         \centering
         \includegraphics[width=\textwidth]{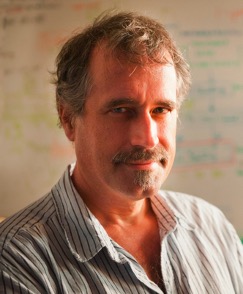}
     \end{subfigure}
     \begin{subfigure}[b]{0.2\textwidth}
         \centering
         \includegraphics[width=\textwidth]{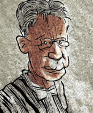}
     \end{subfigure}
     \hfill
     \begin{subfigure}[b]{0.2\textwidth}
         \centering
         \includegraphics[width=\textwidth]{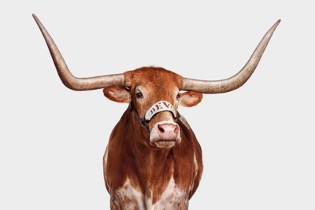}
     \end{subfigure}
          \begin{subfigure}[b]{0.2\textwidth}
         \centering
         \includegraphics[width=\textwidth]{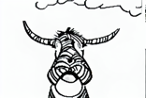}
     \end{subfigure}
     
    \caption{\small \textbf{Left} Input images of Jay Hartzell, Alan Bovik, and the UT Austin mascot Bevo. \textbf{Right} Results of our fine-tuned stable diffusion model performing style transfer on these images into the Calvin and Hobbes comics style.}
    \label{fig:Teaser}
\end{figure}
}

\newcommand{\figForwardDiffusion}{
\begin{figure}
    \centering
    \includegraphics[width=\linewidth]{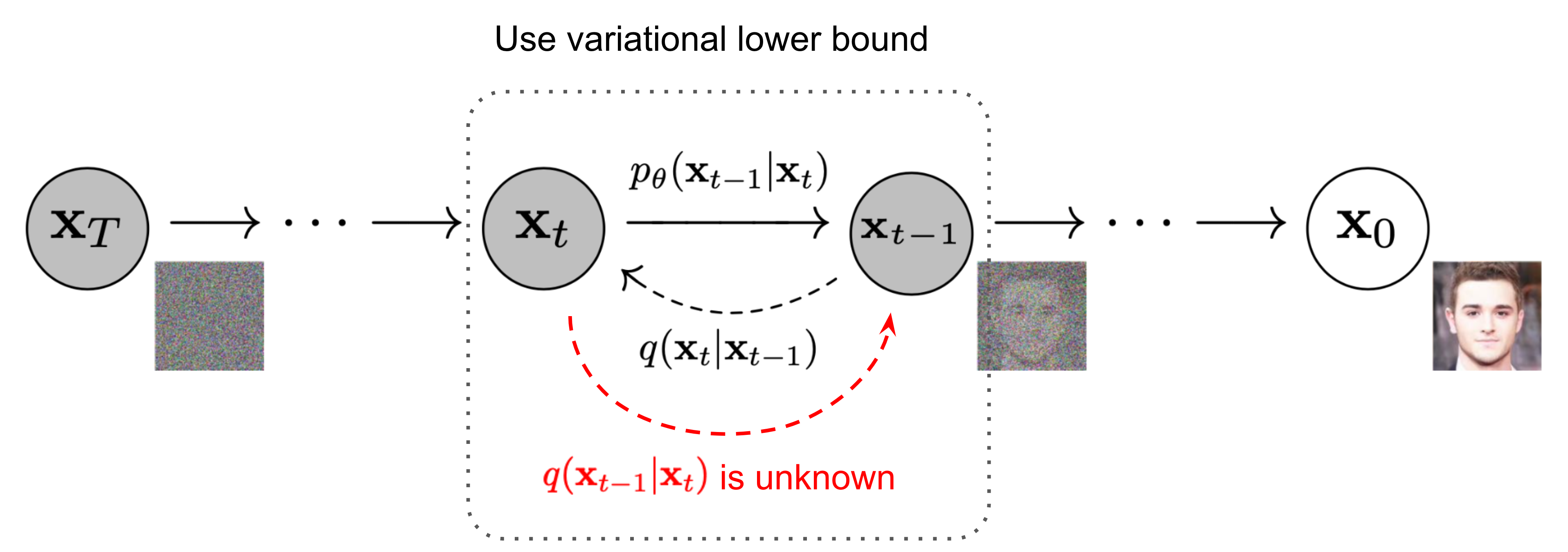}
    \caption{\small The Markov chain of forward (reverse) diffusion process of generating a sample by slowly adding (removing) noise.}
    \label{fig:ForwardDiffusion}
\end{figure}
}

\newcommand{\figDiffusionProcess}{
\begin{figure}
    \centering
    \includegraphics[width=\linewidth]{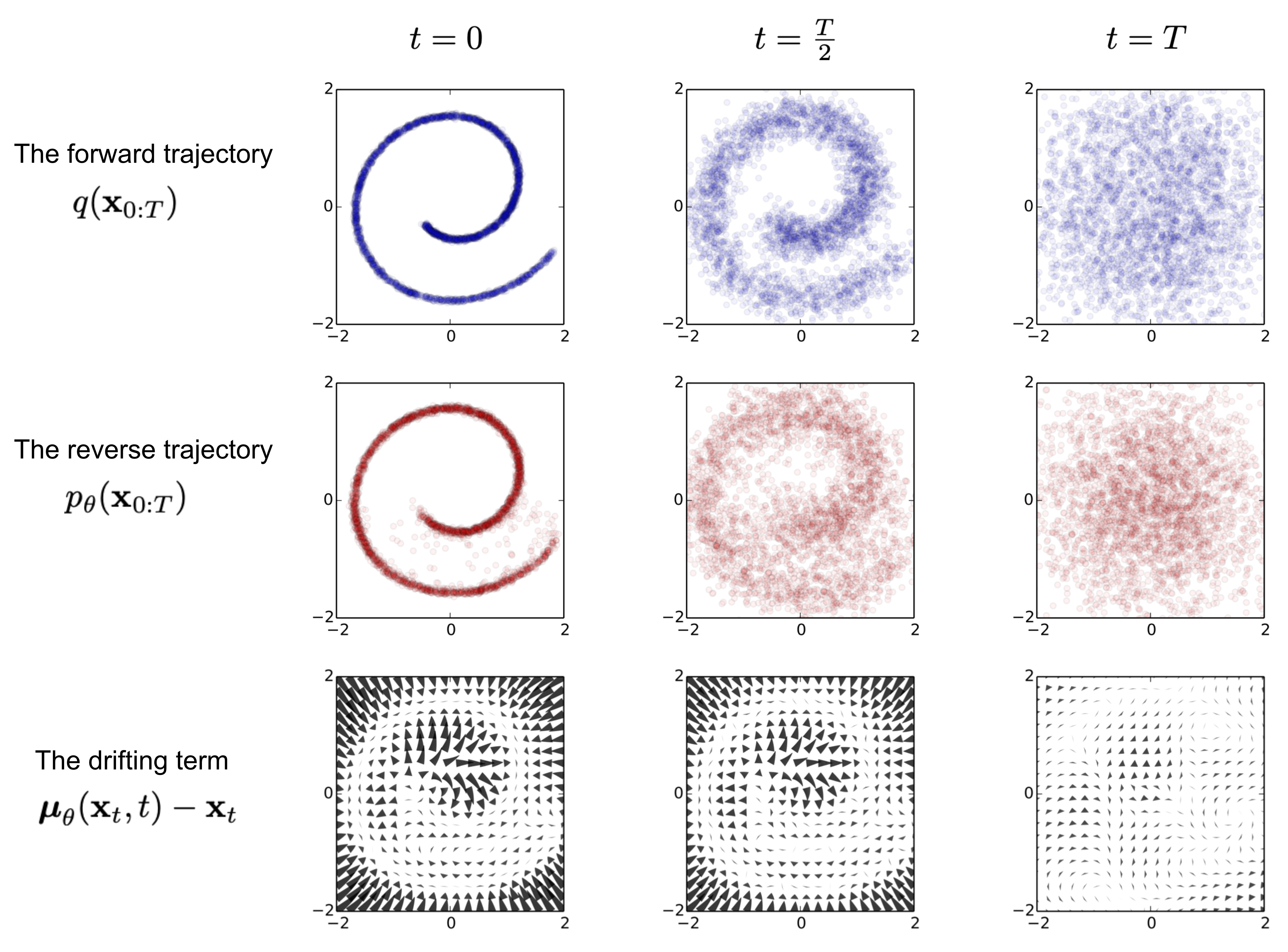}
    \caption{\small An example of training a diffusion model for modeling a 2D swiss roll data.}
    \label{fig:DiffusionProcess}
\end{figure}
}

\newcommand{\figCLIP}{
\begin{figure}
    \centering
    \includegraphics[width=\linewidth]{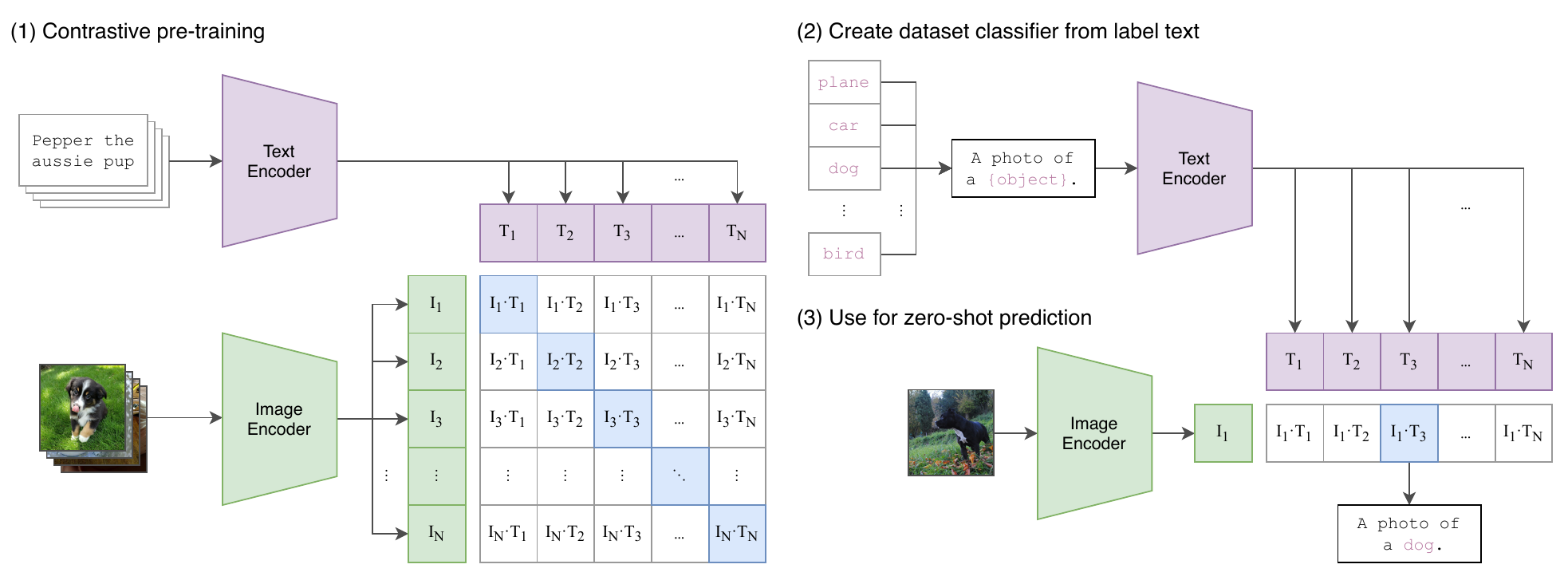}
    \caption{\small Stable diffusion uses CLIP for jointly training an image encoder and a text encoder to predict the correct pairings of $\mathrm{image, text}$}
    \label{fig:CLIP}
\end{figure}
}

\newcommand{\figLatentDiffusionArch}{
\begin{figure}
    \centering
    \includegraphics[width=\linewidth]{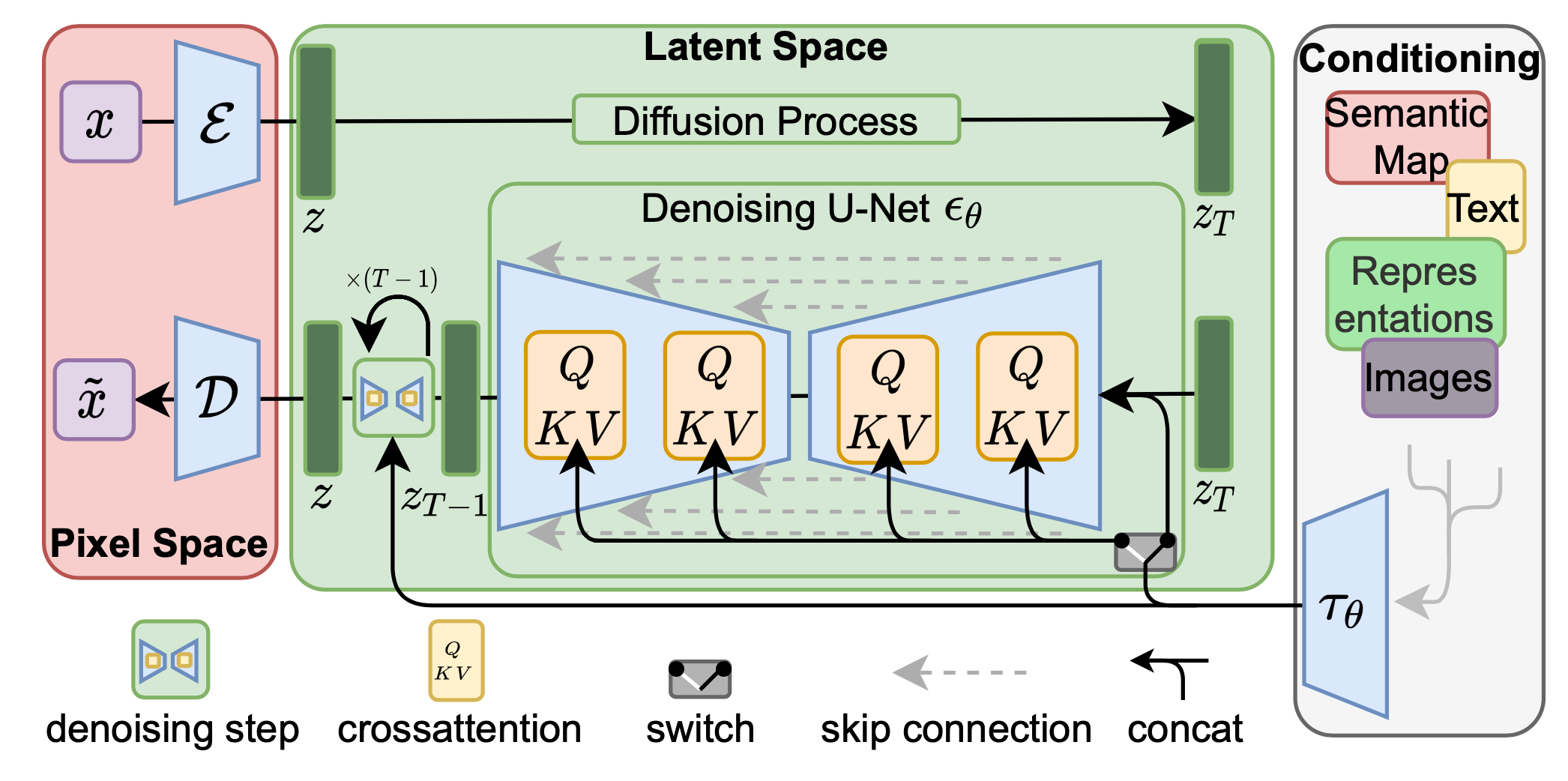}
    \caption{\small The architecture of latent diffusion model.}
    \label{fig:LatentDiffusionArch}
\end{figure}
}

\newcommand{\figTxtImgBeforeFineTuning}{
\begin{figure}[!htbp]
     \centering
     \begin{subfigure}[b]{0.15\textwidth}
         \centering
         \includegraphics[width=\textwidth]{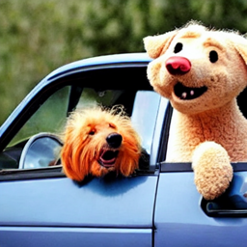}
         \caption{}
     \end{subfigure}
     \begin{subfigure}[b]{0.15\textwidth}
         \centering
         \includegraphics[width=\textwidth]{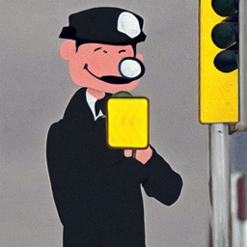}
         \caption{}
     \end{subfigure}
     \begin{subfigure}[b]{0.15\textwidth}
         \centering
         \includegraphics[width=\textwidth]{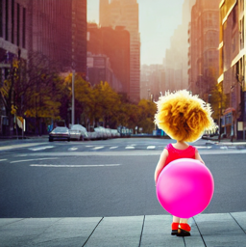}
         \caption{}
     \end{subfigure}
     \caption{Txt2Img Image generation before fine tuning}
     \label{fig:figTxtImgBeforeFineTuning}
\end{figure}
}

\newcommand{\figTxtImgAfterFineTuning}{
\begin{figure}[!htbp]
     \centering
     \begin{subfigure}[b]{0.15\textwidth}
         \centering
         \includegraphics[width=\textwidth]{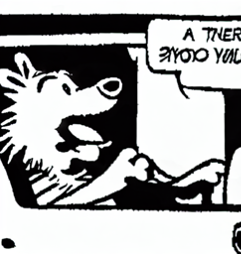}
         \caption{}
     \end{subfigure}
     \begin{subfigure}[b]{0.15\textwidth}
         \centering
         \includegraphics[width=\textwidth]{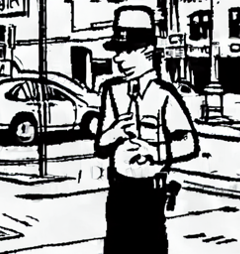}
         \caption{}
     \end{subfigure}
     \begin{subfigure}[b]{0.15\textwidth}
         \centering
         \includegraphics[width=\textwidth]{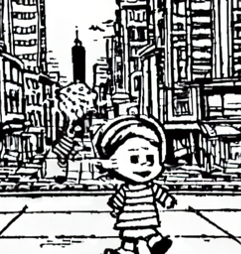}
         \caption{}
     \end{subfigure}
     \caption{Txt2Img Image generation after fine tuning}
     \label{fig:figTxtImgAfterFineTuning}
\end{figure}
}

\newcommand{\imagescalar}{0.19}

\newcommand{\figResults}{

\begin{figure*}[!htbp]
     \centering
     \begin{subfigure}[b]{\imagescalar\textwidth}
         \centering
         \includegraphics[width=\textwidth]{assets/img2img_jay_hartzel.jpg}
     \end{subfigure}
     \begin{subfigure}[b]{\imagescalar\textwidth}
         \centering
         \includegraphics[width=\textwidth]{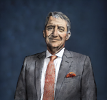}
     \end{subfigure}
     \begin{subfigure}[b]{\imagescalar\textwidth}
         \centering
         \includegraphics[width=\textwidth]{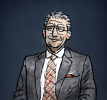}
     \end{subfigure}
     \begin{subfigure}[b]{\imagescalar\textwidth}
         \centering
         \includegraphics[width=\textwidth]{assets/img2img_jay_hartzel2.png}
     \end{subfigure}
     \begin{subfigure}[b]{\imagescalar\textwidth}
         \centering
         \includegraphics[width=\textwidth]{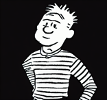}
     \end{subfigure}
     \label{fig:figImgImgOrigiJay}

     \begin{subfigure}[b]{\imagescalar\textwidth}
         \centering
         \includegraphics[width=\textwidth]{assets/img2img_bovik.jpg}
     \end{subfigure}
     \begin{subfigure}[b]{\imagescalar\textwidth}
         \centering
         \includegraphics[width=\textwidth]{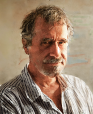}
     \end{subfigure}
     \begin{subfigure}[b]{\imagescalar\textwidth}
         \includegraphics[width=\textwidth]{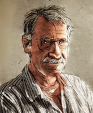}
     \end{subfigure}
     \begin{subfigure}[b]{\imagescalar\textwidth}
         \centering
         \includegraphics[width=\textwidth]{assets/img2img_bovik2.png}
     \end{subfigure}
     \begin{subfigure}[b]{\imagescalar\textwidth}
         \centering
         \includegraphics[width=\textwidth]{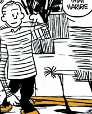}
     \end{subfigure}
     \label{fig:figImgImgOrigiBovik}

     \begin{subfigure}[b]{\imagescalar\textwidth}
         \centering
         \includegraphics[width=\textwidth]{assets/img2img_bevo.jpg}
     \end{subfigure}
     \begin{subfigure}[b]{\imagescalar\textwidth}
         \centering
         \includegraphics[width=\textwidth]{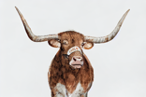}
     \end{subfigure}
     \begin{subfigure}[b]{\imagescalar\textwidth}
         \centering
         \includegraphics[width=\textwidth]{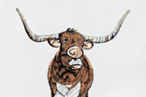}
     \end{subfigure}
     \begin{subfigure}[b]{\imagescalar\textwidth}
         \centering
         \includegraphics[width=\textwidth]{assets/img2img_bevo2.png}
     \end{subfigure}
     \begin{subfigure}[b]{\imagescalar\textwidth}
         \centering
         \includegraphics[width=\textwidth]{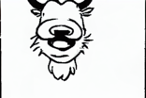}
     \end{subfigure}
     \label{fig:figImgImgOrigiBevo}

     \begin{subfigure}[b]{0.15\textwidth}
         \centering
         \includegraphics[width=\textwidth]{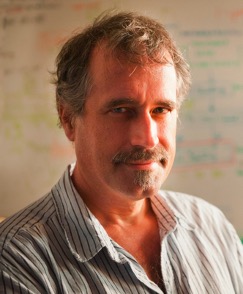}
     \end{subfigure}
     \begin{subfigure}[b]{0.15\textwidth}
         \centering
         \includegraphics[width=\textwidth]{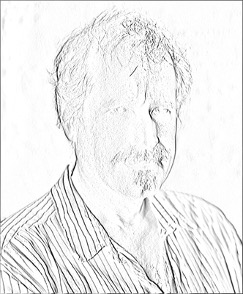}

     \end{subfigure}
     \begin{subfigure}[b]{0.15\textwidth}
         \centering
         \includegraphics[width=\textwidth]{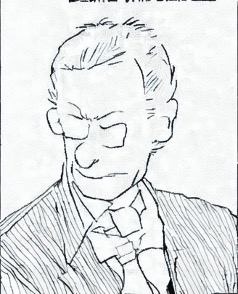}

     \end{subfigure} 
     \label{fig:figImgImgEdgeBovik}
    \hfill
     \begin{subfigure}[b]{0.15\textwidth}
         \centering
         \includegraphics[width=\textwidth]{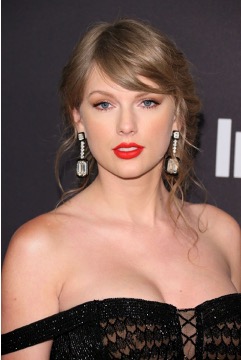}
     \end{subfigure}
     \begin{subfigure}[b]{0.15\textwidth}
         \centering
         \includegraphics[width=\textwidth]{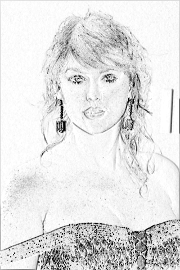}
     \end{subfigure}
     \begin{subfigure}[b]{0.15\textwidth}
         \centering
         \includegraphics[width=\textwidth]{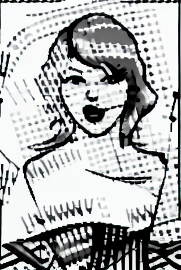}
     \end{subfigure}
     \label{fig:figImgImgEdgeTayTay}

     \begin{subfigure}[b]{\textwidth}
         \centering
         \includegraphics[width=\textwidth]{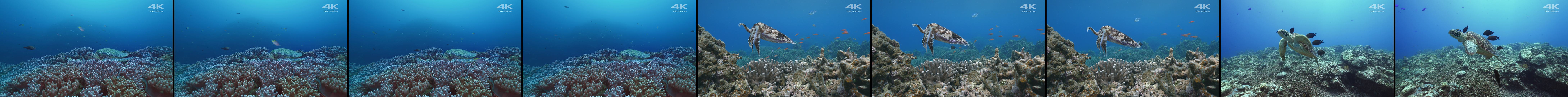}
         \label{fig:figVideo_input}
     \end{subfigure}
     \begin{subfigure}[b]{\textwidth}
         \centering
         \includegraphics[width=\textwidth]{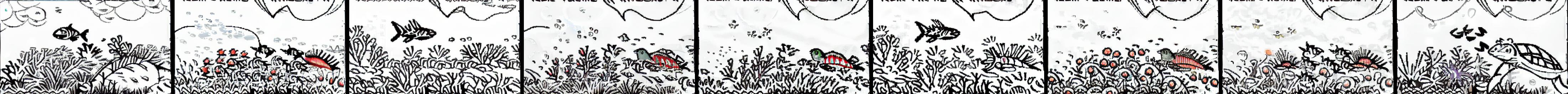}
         \label{fig:figVideo_output}
     \end{subfigure}
    
     \caption{\small Here are all of our results for different input images, as well as different input modalities. \\
     \textbf{Row 1:} Img2Img Image generation with Jay Hartzel’s original image input \\
     \textbf{Row 2:} Img2Img Image generation with Prof. Bovik’s original image input \\
     \textbf{Row 3:} Img2Img Image generation with Bevo’s original image as input \\
     \textbf{Row 4:} Img2Img Image generation with Prof. Bovik’s edeg map input; Img2Img Image generation with Taylor Swift’s edge map input\\
     \textbf{Row 5:} Sequence of 8 frames of the input video \\
     \textbf{Row 6:} Sequence of 8 frames of the output video
    }
    \label{fig:figVideo}
    \end{figure*}
 }

\title{Style transfer using Stable Diffusion}
\name{ Asvin Kumar Venkataramanan, Sloke Shrestha, Sundar Sripada Venugopalaswamy Sriraman}
\address{Department of Electrical and Computer Engineering, \\ The University of Texas at Austin}

\begin{document}

\maketitle

%

%

\begin{abstract}
This project report summarizes our journey to perform stable diffusion fine-tuning on a dataset containing Calvin and Hobbes comics. The purpose is to convert any given input image into the comic style of Calvin and Hobbes, essentially performing style transfer. We train stable-diffusion-v1.5 using Low Rank Adaptation (LoRA) to efficiently speed up the fine-tuning process. The diffusion itself is handled by a Variational Autoencoder (VAE), which is a U-net. Our results were visually appealing for the amount of training time and the quality of input data that went into training.
\end{abstract}


\section{Introduction}
\label{sec:intro}

In the realm of artistic expression and cultural preservation, the desire to revitalize a timeless treasure such as Calvin and Hobbes comics has fueled our motivation to seamlessly blend nostalgia with modern techniques. This endeavor not only aims to breathe new life into the beloved comic strips but also stands as a testament to the ongoing exploration of advanced machine learning approaches.

\figTeaser

One such avenue of exploration is the utilization of stable diffusion fine-tuning, a cutting-edge process that holds the promise of achieving a delicate balance between preserving the essence of the original artwork and infusing it with contemporary flair. By undertaking this project, we endeavor to unravel the intricacies of stable diffusion fine-tuning, delving into its nuances to better comprehend its potential applications in the domain of artistic style transfer.

Amidst a myriad of deep learning based style transfer approaches, we chose stable diffusion as our preferred method due to its unique ability to synthesize artistic styles. The decision to leverage LoRA (Low-Rank Adaptive) for fine-tuning plays a pivotal role in optimizing the training process. LoRA enables a significant acceleration in training speed, unlocking the potential for a more streamlined and effective convergence towards the desired style transfer outcomes. LoRA combined with the underlying diffusion model offers a compelling solution to revitalize Calvin and Hobbes comics.

In summary, this report submitted for the 2023 edition of \textit{ECE371Q Digital Image Processing} by Dr. Bovik describes our attempt at learning how to perform fine-tuning of stable diffusion models and revitalize Calvin and Hobbes comics along the way.

\section{DATASET PREPARATION}
\label{sec:dataset}

To create the dataset for the fine-tuning process, we downloaded 11 Volumes of the Calvin and Hobbes comics from the Internet Archive. These PDF documents were available free of copyright. We then extracted pages from these documents. Pages from the beginning and the end with the foreward and publishing details were discarded since we wanted to focus on images with the original comic strips. 

\subsection{Black and White Pages}
The original Calvin and Hobbes comic strips were syndicated and published in newspapers worldwide every day for 10 years from 1985 to 1995. The comic strips published from Monday to Saturday were in black and white and consisted of four panels in a single row. The comics published on Sundays had 3 rows of images all in color. 

We separated the pages based on color by using simple image processing techniques. The black and white pages are "nearly" binary images as the pixels are clustered in two modes around 3 and 250 in grayscale values. We use this fact to sort the images. If \(I_R, I_G, I_B\) represent the red, green and blue channels of the input image respectively, we analyze the pixelwise difference between two consecutive channels, i.e. \(abs(I_R-I_G)\) and \(abs(I_G-I_B)\). The black and white images have a max absolute channel-wise difference of 10 while this does not hold true for color images as the red, green and blue channels encode different intensities. 

Once we sort the pages based on color, we decided to work solely with black and white comic strips sunce we had nearly twice as many samples for black and white as we had for colored ones. Another motivation for this choise was the fact that the weekday black and white comic strips tend to follow a consistent and reliable four panel structure that allows us to easily extract panels of the same size. The weekend colored comic strips tend to be more artistic and do not follow the same sizes for each panel and thus pose a lot of variation in the dataset. 

\subsection{Panel Extraction}
To extract panels from the black and white images, we used simple coordinate-based cropping techniques. Each page features two rows of comic strips and each comic strip contains 4 panels. We started with 11 Volumes and approximately 166 pages in each volume. Considering we discard a third of the pages because they are in color, we expected approximately \(11*166*2/3*2*4=9738\) panels. Due to variation in number of pages between the volumes, we ended up with 11,033 black and white panels of the same size. 

\subsection{Text Captions}
To create a dataset for fine-tuning diffusion models, each image in the dataset needs a meaningful accompanying text caption. Since these images were manually generated, they were missing any alt-text captions. 

We explored the idea of using the multi-modal vision-language framework BLIP2 \cite{blip2}, capable of answering questions based on any input image, to generate captions for our input panels. However, the generated captions were unsatisfactory. 

Another approach we tried was to use the multi-modal GPT-4 \cite{gpt4} that can accept image and text inputs to generate text outputs. This service provided high quality accurate captions for our image panels. However, GPT-4, being a paid service, proved infeasible too. 

Finally, we settled with using the same caption for all images. Although this is not an ideal choice, we expected the diffusion model to generalize sufficiently owing to the large dataset size. We used the synthetic keyword "CNH3000" to avoid any potential clashes with existing text prompts that the diffusion model is already familiar with. 

\section{Methodology}
\label{sec:method}

As described in Section \ref{sec:intro}, we use a denoising diffusion probabilistic model \cite{ddpm} to perform a style transfer operation. Specifically, we use the open-source model Stable Diffusion v1.5 \cite{sdv15} from RunwayML. To fine-tune this large model efficiently, we use a training technique developed by Microsoft Research called LoRA \cite{lora} which stands for Low Rank Adaptation. More about the network and the training procedure in the following two subsections. 

\subsection{Network}
\label{ssec:network}

The Stable Diffusion Model v1.5 consists of many pieces. (Describe all that here)

\subsubsection{Forward Diffusion Process}
Given a data point sampled from a real data distribution $\mathbf{x}_0 \sim q(\mathbf{x})$, let us define a forward diffusion process in which we add small amount of Gaussian noise to the sample in steps, producing a sequence of noisy samples. The step sizes are controlled by a variance schedule ${\beta_t \in (0,1)}_{t=1}^{T}$.

$q\left(\mathbf{x}_t \mid \mathbf{x}_{t-1}\right)=\mathcal{N}\left(\mathbf{x}_t ; \sqrt{1-\beta_t} \mathbf{x}_{t-1}, \beta_t \mathbf{I}\right) \quad q\left(\mathbf{x}_{1: T} \mid \mathbf{x}_0\right)=\prod_{t=1}^T q\left(\mathbf{x}_t \mid \mathbf{x}_{t-1}\right)$.

The data sample $\mathbf{x}_0$ gradually loses its distinguishable features as the step $t$ becomes larger. Eventually when $T \to\infty, \mathbf{x}_T$ is equivalent to an isotropic Gaussian distribution.

\figForwardDiffusion

\subsubsection{Reverse Diffusion Process}

If we can reverse the above process and sample from $q(\mathbf{x}_{t-1} | \mathbf{x}_{t})$, we will be able to recreate the true sample from a Gaussian noise input, $\mathbf{x}_{T} \sim \mathcal{N}(\mathbf{0},\mathbf{I})$. Note that if $\beta_t$ is small enough, $q(\mathbf{x}_{t-1} | \mathbf{x}_{t})$ will also be Gaussian. Unfortunately, we cannot easily estimate $q(\mathbf{x}_{t-1} | \mathbf{x}_{t})$ because it needs to use the entire dataset and therefore we need to learn a model $p_0$ to approximate these conditional probabilities in order to run the reverse diffusion process.

$p_\theta\left(\mathbf{x}_{0: T}\right)=p\left(\mathbf{x}_T\right) \prod_{t=1}^T p_\theta\left(\mathbf{x}_{t-1} \mid \mathbf{x}_t\right) \quad p_\theta\left(\mathbf{x}_{t-1} \mid \mathbf{x}_t\right)=\mathcal{N}\left(\mathbf{x}_{t-1} ; \boldsymbol{\mu}_\theta\left(\mathbf{x}_t, t\right), \boldsymbol{\Sigma}_\theta\left(\mathbf{x}_t, t\right)\right)$

\figDiffusionProcess

\subsection{Text and Image Encoding}
For text conditioning, we use Contrastive Language-Image Pre-training (CLIP) \cite{radford2021learning}. CLIP embeds text and image in the same space via a projection layer. Thus, it can efficiently learn visual concepts, in the form of text, via natural language supervision and perform zero-shot classification (Figure~\ref{fig:CLIP})

In the pre-training stage, the image and text encoders are trained to predict which images are paired with which texts in a dataset of 400M image-caption pairs. CLIP is trained to maximize the cosine similarity of the image and text embeddings of image-caption pairs via a multi-modal embedding space.

\figCLIP

Latent diffusion model \cite{sdv15} runs the diffusion process in the latent space instead of pixel space, making training cost lower and inference speed faster. It is motivated by the observation that most bits of an image contribute to perceptual details and the semantic and conceptual composition still remains after aggressive compression. LDM loosely decomposes the perceptual compression and semantic compression with generative modeling learning by first trimming off pixel-level redundancy with autoencoder and then manipulate/generate semantic concepts with diffusion process on learned latent.

The perceptual compression process relies on an autoencoder model. An encoder $\mathcal{E}$
is used to compress the input image 
$\mathbf{x} \in \mathbb{R}^{H \times W \times 3}$ to a smaller 2D latent vector $\mathbf{z}=\mathcal{E}(\mathbf{x}) \in \mathbb{R}^{h \times w \times c}$, where the downsampling rate $f=H / h=W / w=2^m, m \in \mathbb{N}$. Then an decoder $\mathcal{D}$ reconstructs the images from the latent vector, $\mathbf{x}^x = \mathcal{D}(z)$. 

The diffusion and denoising processes happen on the latent vector $\mathbf{z}$. The denoising model is a time-conditioned U-Net, augmented with the cross-attention mechanism to handle flexible conditioning information for image generation (e.g. class labels, semantic maps, blurred variants of an image). The design is equivalent to fuse representation of different modality into the model with cross-attention mechanism. Each type of conditioning information is paired with a domain-specific encoder $\tau_\theta$ to project the conditioning input $y$ to an intermediate representation that can be mapped into cross-attention component, $\tau_\theta (y) \in \mathbb{R}^{M \times d_\tau}$.

\figLatentDiffusionArch

\subsection{Training}
\label{ssec:training}
We use a training method called Low Rank Adaptation (LoRA) \cite{lora} to fine-tune the diffusion model. LoRA is a technique proposed by researchers at Microsoft Research to fine-tune Large Language Models. Other researchers have found that this method can be successfully adapted to diffusion models as well. In this technique, weight matrices in existing layers, typically attention layers, are fine-tuned to a specific dataset by adding update matrices. These update matrices are further decomposed as a product of two matrices of lower-rank. During the fine-tuning process, the original weights are frozen and the weights in the update matrices are learnt. The LoRA framework is flexible as we have control over the rank of the matrices in the decomposition of the update matrix. It also allows for using multiple low-rank update matrices simultaneously. 

Consider an attention layer with a weight matrix $ W \in \mathbf{R}^{d \times h}$ then we can associate it with an update matrix $\Delta W = B A $ where $B \in \mathbf{R}^{d \times k}$ and $A \in \mathbf{R}^{k \times h}$. In our experiments, we set $k=4$ and we use one update matrix for each attention layer in the UNet.

Fine-tuning using LoRA has several advantages. First, this approach uses far less memory than traditional fine-tuning approaches since we only need to compute the gradients for the relatively smaller weight matrices. This has the added benefit that the training process is faster. Another challenge in using traditional fine-tuning approaches for diffusion models that LORA overcomes is one called catastrophic forgetting or catastrophic interference. \cite{catforget1} \cite{catforget2} \cite{catforget3} \cite{catforget4} In DDPMs, this manifests as the model forgetting old text-to-image associations upon learning new ones. Since the weight matrices from the base model are frozen, original knowledge is preserved. 

For our custom Calvin and Hobbes dataset, we fine tune the model on 11,000 input images paired with the synthetic text caption "CNH3000" as mentioned earlier. We train with a batch size of 1 and train for 30,000 steps. At each step, a random image from the dataset and a random denoising time step for the DDPM are sampled. These are used to create the noisy and denoised versions of the image at the sampled denoising time step. For example, we might sample image number 1,345 from the dataset and a denoising time step of 33 for training step 13,576/30,000. In that training step, the UNet model receives a time embedding corresponding to 33 and a corresponding noisy and denoised image pair. In this framework, it's easy to see that the denoising network sees approximately 600 pairs of examples for each denoising time step. This number is far less than the 11,000 samples we have. Though the network does not train on all available input images at each time steps, it is able to generalize the denoising process with a significantly smaller subset. This is also important as we do not want our diffusion model to overfit to the available data. 

The network was fine-tuned with an initial learning rate of 1e-4. We used a cosine learning rate scheduler with restarts to decrease the learning rate gradually to 0 over a period of 15,000 training steps. Overall, the fine-tuning process took about 6 hours for our choice of hyperparameters while providing satisfactory results.

\section{Experiments \& Results}
\label{sec:results}

We performed four major experiments: Text to Image, Image to Image with original image, Image to Image with edge map inputs, and Videos.

\subsection{Text to Image}

\figTxtImgBeforeFineTuning

First, we observe the images generated before fine tuning stable diffusion on Calvin and Hobbes images. Figure \ref{fig:figTxtImgBeforeFineTuning} shows the images generated without fine tuning. The text prompts given were:

\begin{enumerate}
    \item A happy dog in car with its head out of the window in the style of Calvin and Hobbes 
    \item A policeman at a traffic light on a busy street wearing a hat in the style of Calvin and Hobbes 
    \item A little girl with a balloon walking down a street with buildings in the background in the style of Calvin and Hobbes
\end{enumerate}

After fine tuning stable diffusion on Calvin and Hobbes images, we generated some more images using the same prompts. The only thing different with the prompt is that we replaced "Calvin and Hobbes" with our keyword, "CNH3000". Figure \ref{fig:figTxtImgAfterFineTuning} shows the images generate after fine tuning. The text prompts given were: 
\begin{enumerate}
    \item A happy dog in car with its head out of the window in the style of CNH3000
    \item A policeman at a traffic light on a busy street wearing a hat in the style of CNH3000
    \item A little girl with a balloon walking down a street with buildings in the background in the style of CNH3000
\end{enumerate}

\figTxtImgAfterFineTuning

\figResults

\label{ssec:t2i}

\subsection{Image to Image}
Stable diffusion takes in noisy images as starting point. Then, the denoising U-Net iteratively removes the noise from the image to generate another image. For text to image models, the denoising process starts with gaussian image sample. For image to image models, the denoising process starts with an image with some noise added to it. This pipeline where we add some noise to the image and start the reverse diffusion process is called the Image2Image pipeline. The noised image with some prompts can allow us to style transfer an image. 

For all these image generation, we used similar prompts as \ref{ssec:t2i} with the keyword, "CNH3000."
\label{ssec:i2i}

\subsection{Image to Image (Edge Map)}
We experimented with starting with noisy sample of edge maps of images instead of using the original image. We hypothesized that this would give a simpler input to the diffusion model which would help the diffusion model produce more cartoon like images.

\label{ssec:i2ie}

\subsection{Videos}
We tried to apply diffusion models' output to individual frames of a video. The last two rows of Figure \ref{fig:figVideo} shows the 8 input and output frames. Notice the inconsistency in the results. The outputs are not temporally cohesive. 
\label{ssec:v2v}


\section{Future Work}

Our work was a relatively simple approach to fine-tuning a baseline diffusion model to the style transfer task. There are many areas with scope for improvement and exploration. 

We could start off with improving the dataset. Currently, we simply extract panels from the original comics. Since these comics aim to convey a short story within 4 panels, they are often filled with text. The generative model learns that these texts are part of the calvin and hobbes style. This is unideal and does not align with a regular person's expectation of the comic's style. Removing the text using available open-source OCR tools like pytesseract is a good starting point. Alternatively, we could generate better captions for the images use tools like GPT-4 or Flamingo \cite{flamingo} to incorporate the text from the images into the captions. This could aid the diffusion model in understanding the style of the comics better. Towards improving the dataset, it would also be interesting to include color images and have the model jointly learn the style from black and white as well as colored panels simultaneously. 

Thinking about the training mechanisms, we explored LoRA as described in Section \ref{ssec:training}. It would be interesting to explore ideas such as DreamBooth \cite{dreambooth}, Textual Inversion \cite{ti} and ControlNet \cite{controlnet} since they have shown promising results for other image-based tasks. 

We briefly experimented with applying our style transfer model on videos as described in Section \ref{ssec:v2v}. Our results were temporaly inconsistent. Exploring more complex models like FFNeRV \cite{ffnerv} and InstructPix2Pix \cite{instructpix2pix} as demonstrated by other projects would be interesting. 

Finally, one of the natural questions in the age of Large Language Models (LLMs) is whether we can use a model to generate entire comic strips in the style of Calvin and Hobbes. Currently, we restrict ourself to style transfer of separate images. Combining LLMs along with a diffusion pipeline to generate consistent and meaningful stories would be an interesting task.

\section{Contribution}
\label{sec:contrib}

All the authors of this paper contributed actively at all stages of the project. If we had to assign credit to specific authors, we would say that Sundar was heavily responsible for dataset creation and pre-processing, Asvin was responsible for the fine-tuning of diffusion models and Sloke was responsible for running extensive experiments with the fine-tuned model.

\section{Acknowledgements}

The authors acknowledge the Texas Advanced Computing Center (TACC) at The University of Texas at Austin for providing HPC resources that have contributed to the results reported within this paper. URL: http://www.tacc.utexas.edu. We would also like to thank Prof. Alan C. Bovik for his insights and guidance in the development of this project.

\bibliographystyle{IEEEbib}
\bibliography{strings,refs}

\end{document}